\documentclass[lettersize,journal]{IEEEtran}

\usepackage{amsmath,amsfonts}
\usepackage{url}
\usepackage{graphicx}
\usepackage{balance}
\usepackage{xcolor}
\usepackage{cite}
\usepackage{bm} 
\usepackage{mathtools}
\usepackage[hidelinks,
			colorlinks=true,
			linkcolor=blue,
			urlcolor=blue,
			citecolor=blue]{hyperref}
\usepackage{todonotes}
\usepackage{tikz}
\usepackage{tabularx}
	
	\newcolumntype{M}[1]{>{\centering\arraybackslash}m{#1}}
	\newcolumntype{C}{>{\centering\arraybackslash}X}
\usepackage{multirow}
\usetikzlibrary{arrows.meta, calc, backgrounds}
\usepackage[inline]{enumitem}
\usepackage{subcaption}
\usepackage{cleveref}
\newcommand{\refSec}[1]{Sec.~\ref{#1}}
\newcommand{\refFig}[1]{Fig.~\ref{#1}}

\newcommand{\refTab}[1]{Table~\ref{#1}}

\Crefname{figure}{Fig.}{Figs.}

\let\origtodo\todo
\newcommand{\basetodo}{\origtodo}

\renewcommand{\todo}[1]{\basetodo[inline]{#1}}

\newcommand{\blue}[1]{{\color{blue}{#1}}}

\newcommand{\jointConfTime}[0]{
	{\bm{q}\left( t \right)}
}

\newcommand{\inR}[1]{
	\in\mathbb{R}^{#1}
}

\newcommand{\subsetOfR}[1]{
	\subset\mathbb{R}^{#1}
}

\newcommand{\volume}[1]{
	{V_{#1}}
}

\newcommand{\volumeWithDepArg}[2]{
	{V_{#1}\left({#2}\right)}
}

\newcommand{\decimatedVolume}[1]{
	{\widetilde{V}_{#1}}
}

\newcommand{\transfMat}[1]{
	{\bm{T}\left({#1}\right)}
}

\newcommand{\transfMatLinkTime}[1]{
	{\bm{T}_{#1}\left( t \right)}
}

\newcommand{\numLinks}{
	{N}
}

\newcommand{\timeInterval}[2]{
	{\left[#1,\thinspace #2\right]}
}

\newcommand{\timeInSec}[1]{
	{#1\,\text{s}}
}

\newcommand{\timeInMin}[1]{
	{#1\,\text{min}}
}

\newcommand{\distInCm}[1]{
	{#1\,\text{cm}}
}

\newcommand{\volMeasure}[1]{
	{#1\,\text{m}^3}
}

\makeatletter
\newcommand\thefontsize{\f@size pt}
\makeatother

\newif\ifnewon
\newcommand{\new}[1]{\ifnewon\blue{#1}\else#1\fi}

\newonfalse 

\begin{document}

\title{Robot Cell Modeling via\\Exploratory Robot Motions}
\author{
	Gaetano~Meli and Niels~Dehio%
	\thanks{
			The authors are with 
			KUKA Deutschland GmbH, 
			Augsburg, 
			Germany.
			}
	\thanks{
			© 2025 IEEE.  Personal use of this material is permitted.  Permission from IEEE must be obtained for all other uses, in any current or future media, including reprinting/republishing this material for advertising or promotional purposes, creating new collective works, for resale or redistribution to servers or lists, or reuse of any copyrighted component of this work in other works.
	}
}


\maketitle

\begin{abstract}
Generating a collision-free robot motion is crucial for safe applications in real-world settings.
This requires an accurate model of all obstacle shapes within the constrained robot cell, 
which is particularly challenging and time-consuming. 
The difficulty is heightened in flexible production lines,
where the environment model must be updated each time the robot cell is modified.
Furthermore, 
sensor-based methods often necessitate costly hardware and calibration procedures, 
and can be influenced by environmental factors 
(e.g., light conditions or reflections).
To address these challenges,
we present a novel data-driven approach to modeling a cluttered workspace, 
leveraging solely the robot’s internal joint encoders to capture exploratory motions.
By computing the corresponding swept volume,
we generate a (conservative) mesh of the environment that is subsequently used for collision checking 
within established path planning and control methods.
Our method significantly reduces the complexity and cost of classical environment modeling
by removing the need for CAD files and external sensors.
We validate the approach with the KUKA LBR iisy collaborative robot in a pick-and-place scenario.
In less than three minutes of exploratory robot motions
and less than four additional minutes of computation time, 
we obtain an accurate model that enables collision-free motions. 
Our approach is intuitive, 
easy-to-use, 
making it accessible to users without specialized technical knowledge.
It is applicable to all types of industrial robots or cobots. 
\end{abstract}


\section{Introduction}
\IEEEPARstart{A}{n} accurate environment model 
is paramount for successfully deploying and operating robot systems without compromising hardware integrity. 
The process of environment modeling involves creating a (digital) representation of the physical world. 
Typically, 
it encompasses data acquisition through (expensive) sensors~\cite{campos2021orb} and integrating this data into coherent models~\cite{tateno2017cnn,bloesch2018codeslam}. 
The output of this process may be a dense point cloud, 
a 3D map, 
or 3D meshes~\cite{mildenhall2021nerf,kerbl20233d}.
Alternatively,
objects in the robot cell are modeled separately through Computer-Aided Design (CAD) files,
incorporating shape and position information.
This approach can yield imprecise results due to the sim-to-real gap. 
Moreover, 
necessary data are often unavailable.
The challenge intensifies in robot cells frequently modified to meet changing production needs.
Based on our experience, 
currently, 
many enterprises avoid modeling the robot surroundings due to the time-consuming, 
expensive, 
and complex nature of the task. 
However, 
this hinders the implementation of applications involving autonomous robots that generate collision-free paths and adapt motions in real-time. 
Instead, 
operators program sub-optimal trajectories offline,
leading to inefficiencies in execution.

\begin{figure}[!t]
	\centering
	\includegraphics[width=0.99\linewidth]{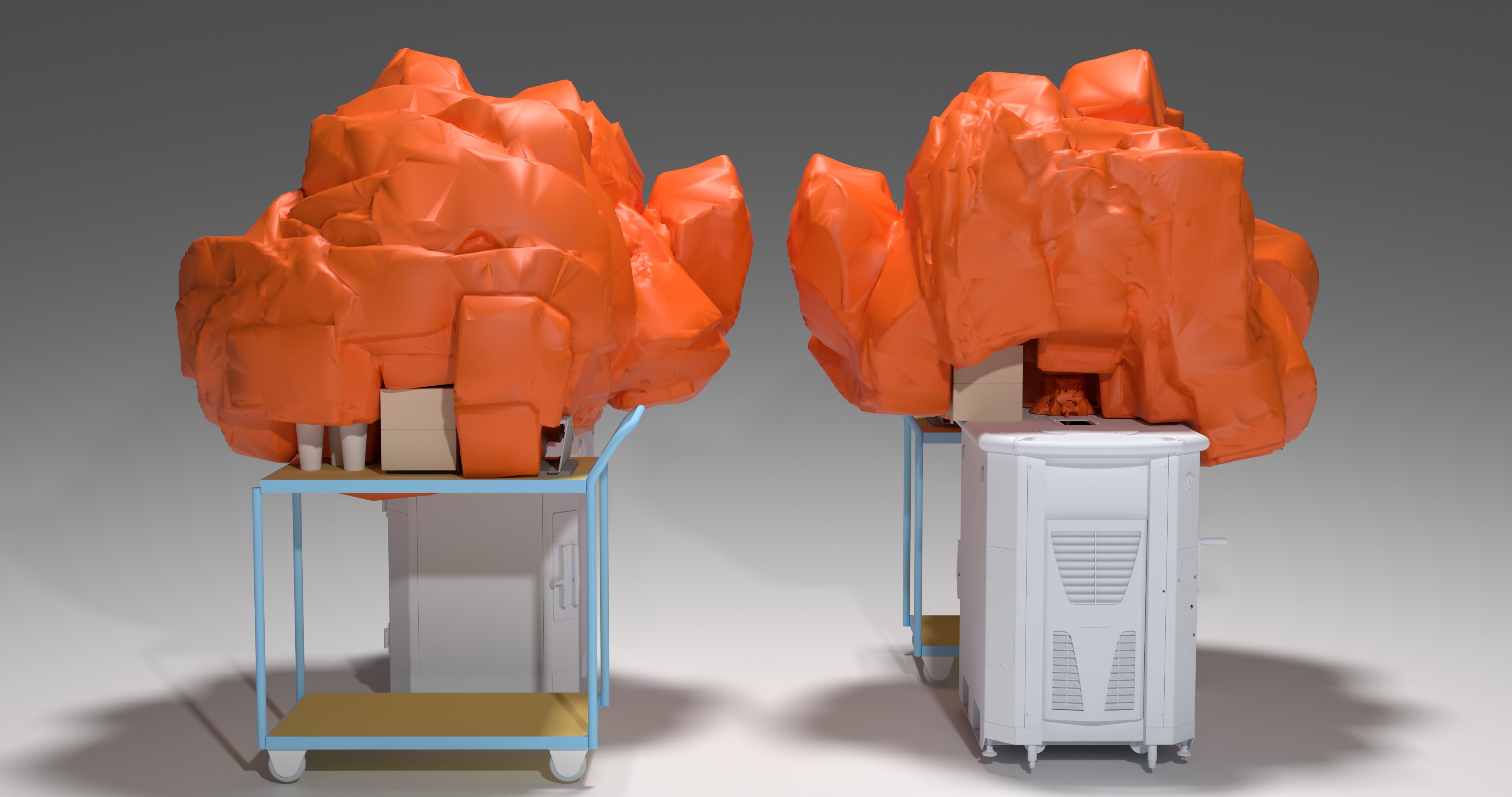}
	\caption{Swept volume (in orange) of a KUKA LBR iisy robot \new{(left: front view, right: back view)}. The exploration of the free workspace has been performed through hand guidance with a cube-shaped exploration tool.}
	\label{fig:sv_pick_and_place}
\end{figure}

\new{Our main contribution is a novel data-driven approach to modeling a 
	static constrained robot cell using exploratory robot motions,
	relying \emph{solely} on the integrated joint encoders. 
	Based on the collected data, 
	we compute the robot swept volume, 
	a non-convex hull represented as a 3D mesh that encapsulates the explored obstacle-free space, 
	focusing on fixed obstacles.
	This mesh represents a conservative approximation of the accessible space. 
	It can be utilized to verify if a particular robot configuration is safe.
	Hence, it can be integrated into state-of-the-art frameworks
	to enable automatic, collision-free motion planning and control~\cite{osorio2020unilateral, fiore2023general}. 
	The assumption of a static environment is valid for many industrial applications, 
	such as welding, painting, gluing, milling, deburring, and inspection. 
	Our approach offers a compelling alternative to CAD-based modeling by eliminating sim-to-real gaps 
	and reducing the need for complex modeling tasks. 
	Moreover, it is fast, cost-effective, and straightforward 
	-- making it well-suited for supporting flexible production lines in modern Industry~4.0 scenarios.
} 
We validate the effectiveness of our approach in a pick-and-place scenario with the KUKA LBR iisy \new{
	(see \refFig{fig:sv_pick_and_place}),
	highlighting the intuitive, user-friendly interface that does not require advanced technical knowledge in robotics.
} 
Note that the approach is robot-agnostic, 
making it applicable to both collaborative and industrial robots.

As a secondary contribution, 
we propose integrating an \textit{exploration tool} that simplifies and accelerates the exploration phase, 
while reducing computational load. 
This tool is inexpensive as it contains no electronics. 
When combined with a commercially available tool change system, 
multiple exploration tools with different shapes can be easily used, 
without significantly extending the exploration process.

A demonstration video of our work is provided as supplementary material and is available at \url{https://youtu.be/p5DA41EtNUA}.

In the following,
\refSec{sec:background} presents the state of the art related to environment modeling and swept volume computation. 
\refSec{sec:freeWorkspaceExploration} describes our approach to obtaining a representation of the \new{obstacle-free} space within the constrained robot cell. 
\new{It also addresses the limitations of our method and potential resolution strategies to overcome them.} 
Experimental results are reported in \refSec{sec:experiments}, 
while \refSec{sec:conclusion} concludes. 

\section{Related work}
\label{sec:background}
\noindent At its core, 
our approach to \emph{environment modeling} is based on the \emph{swept volume} of an exploratory robot motion. 
This section reviews the current state of the art in these two areas.

\subsection{Environment Modeling}
To ensure autonomous robots can safely execute tasks and avoid unintended collisions with their surroundings, 
an accurate environment model is essential.
\emph{Simultaneous Localization and Mapping} (SLAM) is a well-established paradigm 
that allows navigating through an unknown environment, 
while simultaneously localizing the robot pose thanks to onboard sensors such as
cameras, 
laser scanners, 
GPSs, 
sonars, 
or LiDARs.
The algorithm proposed in~\cite{campos2021orb} has proven to be robust, 
accurate, 
and flexible in many applications with various sensor setups. 
However, it faces challenges maintaining high performance within sparsely textured and dynamic environments. 
Remarkable progress in camera localization and map reconstruction has been achieved by integrating deep learning techniques, 
considerably improving the underlying feature extraction~\cite{tateno2017cnn, bloesch2018codeslam}. 
In recent years, 
there has been a growing need for continuous surface modeling and finding solutions for occlusions and sparse observations. 
This has increased research interest in Neural Radiance Fields (NeRF)~\cite{mildenhall2021nerf} and 3D Gaussian Splatting (3DGS)~\cite{kerbl20233d}. 
These methods can produce 3D meshes from dense and compact environment maps. 
However, 
their practical applications may be hindered by limitations in real-time processing, 
hardware demands, 
and training duration.
In addition to these automatic methods, 
CAD models describe the geometry of individual objects in a scene.
They are also used to visualize, 
simulate and optimize large production line processes in 3D,
demanding significant technical expertise.
Although this approach can provide high precision and control when performed accurately,
it may suffer from sim-to-real-gaps and is time-consuming.

\subsection{Swept Volume}
Swept volume (SV) refers to the three-dimensional space encompassing all points that a rigid object motion occupies.
This concept is now widely utilized in several application fields,
including numerically controlled machining verification (e.g., for a milling process), 
modeling of complex solids, 
robot reachable and dexterous workspace analysis, 
collision detection/avoidance, 
and ergonomics.

Abdel-Malek et al.~\cite{abdel2006swept} compared several methods for SV computation.
An \textit{explicit} representation, based on the geometric properties of the moving object, 
is typically obtained via voxel grid approximations 
or by using a triangle mesh to approximate the boundary of the SV. 
This approach struggles to generalize to all motions and object types, 
and the error, 
closely tied to computational power, 
is difficult to control.
Alternatively,
an \textit{implicit} representation describes a mathematical function determining whether a point is inside or outside the SV.
Even though the mathematical formulation is straightforward, 
the SV computation may result in a relevant computational load and provide sub-optimal solutions.
Such numerical issues are avoided by the \textit{stamping} method, which samples the object's motion in space and time.  
The accuracy of the final result heavily depends on the complexity of the object's motion and on the sampling time. 
Moreover, 
it scales poorly with the volume size. 
To the best of our knowledge, 
Sellán et al.~\cite{sellan2021swept} describe the current best-performing method for SV computation by combining the implicit representation with a numerical continuation method.

In robotics, swept volumes have been employed for collision detection and collision-free path planning. 
To ensure safe motions,
\cite{taubig2011real} checked pairwise self-collisions for all robot links utilizing swept convex hulls extended by a buffer radius. 
Baxter et al.~\cite{baxter2020deep} introduced a neural network that predicts the SV geometry for a robot moving from a start to goal joint configuration.
The method outputs discretized voxel grids, 
where each voxel indicates either free or swept space. 
In the same scenario,
\cite{joho2024neural} overcame the accuracy limitation given by the voxel discretization 
by learning a neural implicit SV model as a signed distance function, 
requiring large amounts of training data associated with a desired motion type.
This approach, 
however, 
does not apply to hand guidance, 
tele-operation or any other human-guided motions.

We are not aware of any prior work that utilize SV in the context of environment modeling.

\section{Free Workspace Exploration}
\label{sec:freeWorkspaceExploration}
\noindent This section describes the entire pipeline for modeling the constraints imposed by static obstacles in the robot cell environment.
\new{
	The proposed method is entirely based on data acquired through exploratory motions, typically lasting only a few minutes, and does not require any additional \emph{external} sensor.
}
Our pipeline consists of four main steps (see \refFig{fig:pipeline}), 
which will be detailed in the following subsections: 
1)~sweep through free space,
2)~swept volume,
3)~volume decimation, and
4)~obstacle representation.
The outcome is a mesh describing the boundary of the explored collision-free space. 
\new{
	Steps 1\,--\,3 can optionally be repeated to explore additional free space and enhance the resulting model.
	
	It is important to note that, similarly to CAD-based environment modeling, 
	our approach does not address self-collisions or collision-free motion planning. 
	Instead, it provides an environment model of the robot cell that can be integrated into established trajectory planning and control frameworks 
	to generate safe robot motions.
}

\begin{figure*}[!t]
	\centering
\tikzset{
	rectShape/.style={line width=0.5mm, rectangle, minimum height=1cm, text width=3.0cm, align=center, font=\fontsize{8}{12}\selectfont},
	arrow/.style={midway, above,font=\fontsize{8}{12}\selectfont}	
}

\begin{tikzpicture}[node distance=4.0cm] 
	\node[rectShape, draw=orange] (rect1) {Sweep through free space\\ (\refSec{subsec:sweepFreeSpace})};
	\node[rectShape, draw=orange, right of=rect1] (rect2) {Swept volume\\ (\refSec{subsec:sweptVolume})};
	\node[rectShape, draw=orange, dashed, right of=rect2] (rect3) {Volume decimation\\ (\refSec{subsec:volumeDecimation})};
	\node[rectShape, draw=orange, right of=rect3] (rect4) {Obstacle representation\\ (\refSec{subsec:obstacleRepresentation})};
	
	\draw[-Stealth, line width=0.35mm] (rect1) -- (rect2) node[arrow] {$\boldsymbol{q}(t)$};
	\draw[-Stealth, line width=0.35mm] (rect2) -- (rect3) node[arrow] {$V_{i}$};
	\draw[-Stealth, line width=0.35mm] (rect3) -- (rect4) node[arrow] {$\widetilde{V}_{i}$};
	\draw[-Stealth, line width=0.35mm] (rect4.east) -- ++(0.740,0) node[arrow]{$V_{O}$};
	
		
	
	
\end{tikzpicture}
	\caption{The proposed pipeline consists of 4 steps. 
	First, exploratory robot motions 
	sweeps the free space of the constrained robot cell. 
	The recorded joint trajectories $\jointConfTime$ are utilized to compute the robot link poses and, thus, 
	the corresponding link swept volumes $\volume{i}$. 
	These 3D meshes can optionally be decimated to obtain a simplified volume $\decimatedVolume{i}$ while preserving the overall shape. 
	We obtain a representation of the unexplored and potentially occupied space $\volume{O}$
	by carving out the link swept volumes from a bounding volume that covers the entire robot workspace.
	Steps 1\,--\,3 can be optionally repeated in additional exploration sessions to improve the representation of $\volume{O}$. 
	The 3D mesh associated with $\volume{O}$ can subsequently be utilized 
	within established methods for a collision-free trajectory planning and control.
	}
	\label{fig:pipeline}
\end{figure*}

\textbf{Assumptions.}
\new{
	The approach involves a fixed-based robot with $\numLinks$ non-static rigid links interconnected by joints,
	and is based on the following assumptions:
	\begin{enumerate}[label=A\arabic*., leftmargin=2.5em]
		\item The robot cell is static,
		i.e., 
		there are no moving obstacles or humans within the workspace;
		\item Joint encoders are calibrated, 
		and the configuration-dependent position and orientation of all links are accessible through precise forward kinematics;
		\item Accurate meshes for all robot links and tools are available.
	\end{enumerate}

	The pipeline steps will be illustrated throughout this section through an ideal planar robot manipulator with $\numLinks=3$ non-static links 
	and a KUKA LBR iisy collaborative robot. 
}


\subsection{Sweep through Free Space} 
\label{subsec:sweepFreeSpace}
The novel idea is to utilize the robot swept volume (SV) resulting from exploratory motions 
to identify the collision-free space within the constrained robot cell. 
During the exploratory phase, 
the robot's volume may be intentionally modified, 
i.e., 
by mounting an additional rigid body on the robot flange.
Increasing the robot's overall shape allows for exploring more space in the same amount of time 
or the same space in less time without changing the robot's velocity.
\new
{
	The modification of the robot volume also supports industrial scenarios requiring robot tool changes to satisfy the desired task scenario. 
}

\subsubsection{Exploratory Robot Motion}
The exploration of the constrained workspace can be performed by a human operator through hand guidance, tele-operation, jogging, or other means.
Alternatively,
the robot may also explore its cell autonomously,
i.e., 
it moves (randomly) through free space and reverts its direction of motion upon contact detection.
The robot's velocity during this exploratory phase does not affect the subsequent steps of our pipeline 
as the robot's SV only depends on the joint configurations. 
The recorded joint trajectories $\jointConfTime$ 
of the exploratory motions are continuous and smooth. 

\subsubsection{Exploration Tool}
The end-effector tool(s) required for the actual task scenario can be used in the exploratory phase. 
However, 
the CAD models of such tools are typically very detailed, 
even when simplified, 
thus leading to high computation times and, 
possibly, 
to numerical instabilities during the subsequent steps of our pipeline. 
\new{
	Similar limitations can occur when utilizing a convex decomposition.  
	Moreover, 
	the exploration phase may become time-consuming in the case of small-size end-effector tools.
}
Therefore, 
in the exploratory phase, 
the robot's shape may be modified by mounting an exploration tool on the flange, 
which increases the robot's overall volume and speeds up the exploration process.
\new{
	In this way, 
	the exploration tool, 
	rigidly connected to the robot flange, 
	is considered part of the last link mesh.
}
Such an exploration tool can be customized for the exploratory motion phase.
For its design, 
we recommend simple geometric shapes made of lightweight material without any electronics involved.
Furthermore, 
choosing a form that encloses the tools 
(e.g., gripper or pneumatic suction cup) used later to achieve the actual task may be beneficial.
The dimensions of the exploration tool may also be determined by considering the specific robot cell setup.
If, 
for example, 
narrower gaps are of interest,
the dimensions of the exploration tool should be designed accordingly. 
Moreover, 
the exploratory phase may also involve multiple exploration tools.
Note that modern tool change systems allow users to quickly and flexibly switch between different tools.
In the future, 
robot manufacturers might deliver new manipulators together with a set of such exploration tools that are cheap to produce.

\begin{figure}[t!]
	\centering
	\includegraphics[width=0.99\linewidth]{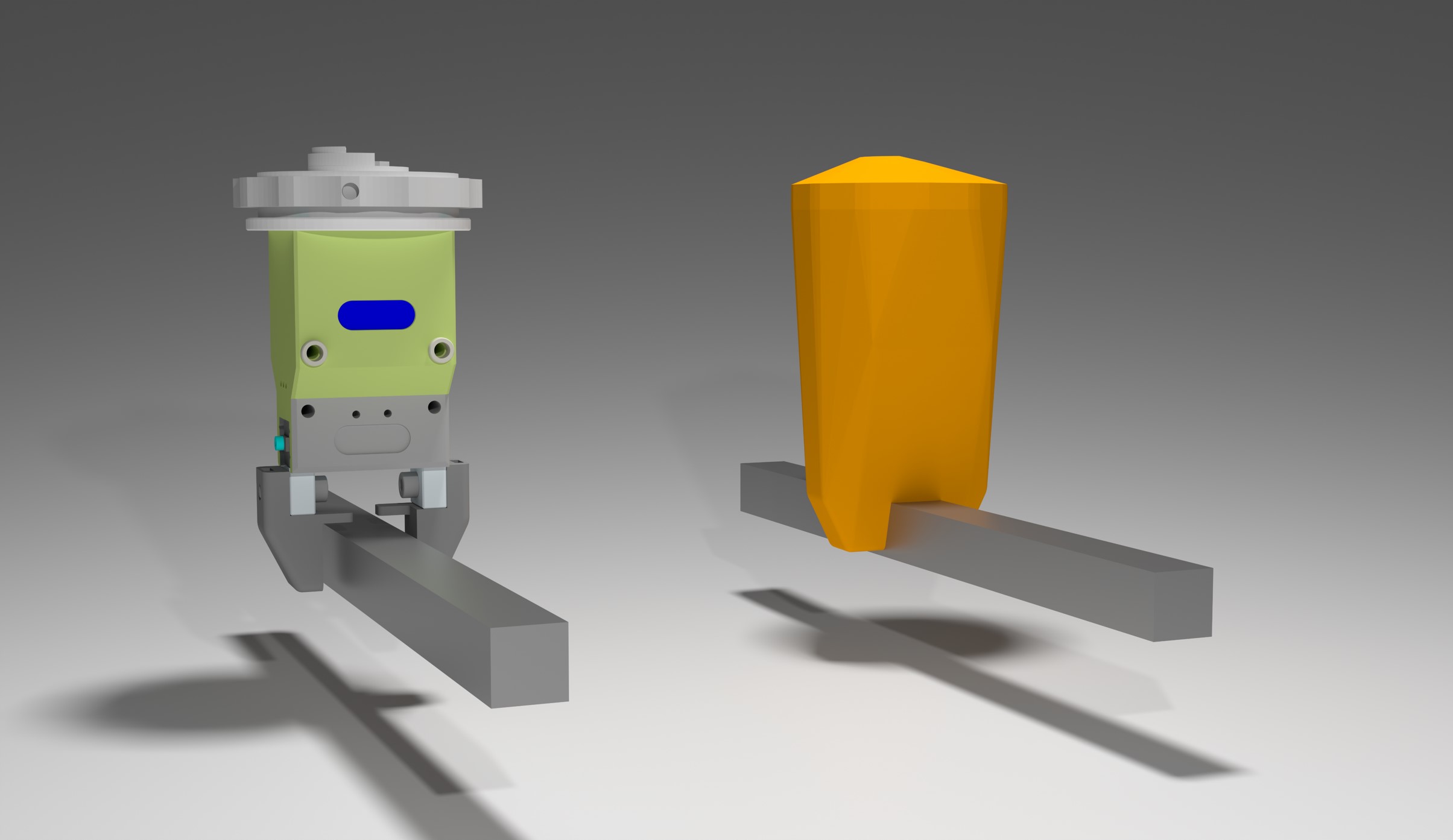}
	\caption{Example scenario where a gripper moves along a static metal bar, positioning its fingers on either side of the obstacle (left). Using the gripper's convex hull (or bounding box) would incorrectly label the volume between the two gripper fingers as obstacle-free, despite being occupied (right).}
	\label{fig:metal_bar}
\end{figure}

\new{
	Utilizing a bounding box or a convex hull to represent the actual tool during the exploration phase constitutes a \emph{non-conservative} approximation within our approach that can lead to severe problems.
	Both representations encompass spatial regions that are not part of the real tool
	-- such as the gap between the two fingers of a parallel jaw gripper.
	Hence, 
	utilizing such representations would erroneously label some regions as obstacle-free, 
	even though they have not been traversed during the exploration phase (see \refFig{fig:metal_bar}). 
	Consequently, 
	this can lead to an incorrect representation of the obstacle-free space, 
	which is unacceptable in safety-critical applications.
	The proposed exploration tool effectively addresses and overcomes this problem.
}

\subsection{Swept Volume} 
\label{subsec:sweptVolume}
The swept volume is the space a rigid body occupies as it moves. 
By definition, 
it encompasses all points within the object at any given moment of the motion.
Given a rigid body~$B$ moving on a path in the time interval $\timeInterval{0}{T}$,
its swept volume~$\volume{S}$ is 
\begin{equation}
	\label{eq:sweptVol}
	\volume{S} = 
	\bigcup_{t\thinspace\in\thinspace\timeInterval{0}{T}}
	\volumeWithDepArg{B}{{\transfMat{t}}},
\end{equation}
where $\volume{B}\subsetOfR{3}$ is the volume occupied by the rigid body,  whose pose along the path is described by the homogeneous transformation matrix $\transfMat{t}$. 

Given the joint trajectories $\jointConfTime$ recorded in the previous step,
the corresponding $i$-th robot link pose $\transfMatLinkTime{i}$ at time~$t$ is obtained 
by utilizing the robot kinematic model.
This allows computing the swept volume $\volume{i}$ of the $i$-th non-static robot link using~\eqref{eq:sweptVol}. 
The computation can be performed for all $\numLinks$ non-static links
\new{-- possibly including the exploration tool as part of the last link --} 
in parallel improving computational efficiency\footnote{Note that the base link of a fixed-base manipulator does not move, 
and, 
hence, 
its swept volume is identical to its body shape.}.
Note that the SV computation is not negatively affected by clustered or heterogeneously distributed data 
(e.g., same or similar joint configurations recorded in the exploration phase).

A volume can generally be represented through a 3D mesh, 
whose faces can be triangles, 
quadrilaterals, or polygons.
In the following, we will consider a triangle mesh to represent a volume without loss of generality.

\begin{figure*}
	\centering
	\begin{minipage}{.49\linewidth}
		\begin{subfigure}[t]{.49\linewidth}
			\centering
			\includegraphics[width=\textwidth]{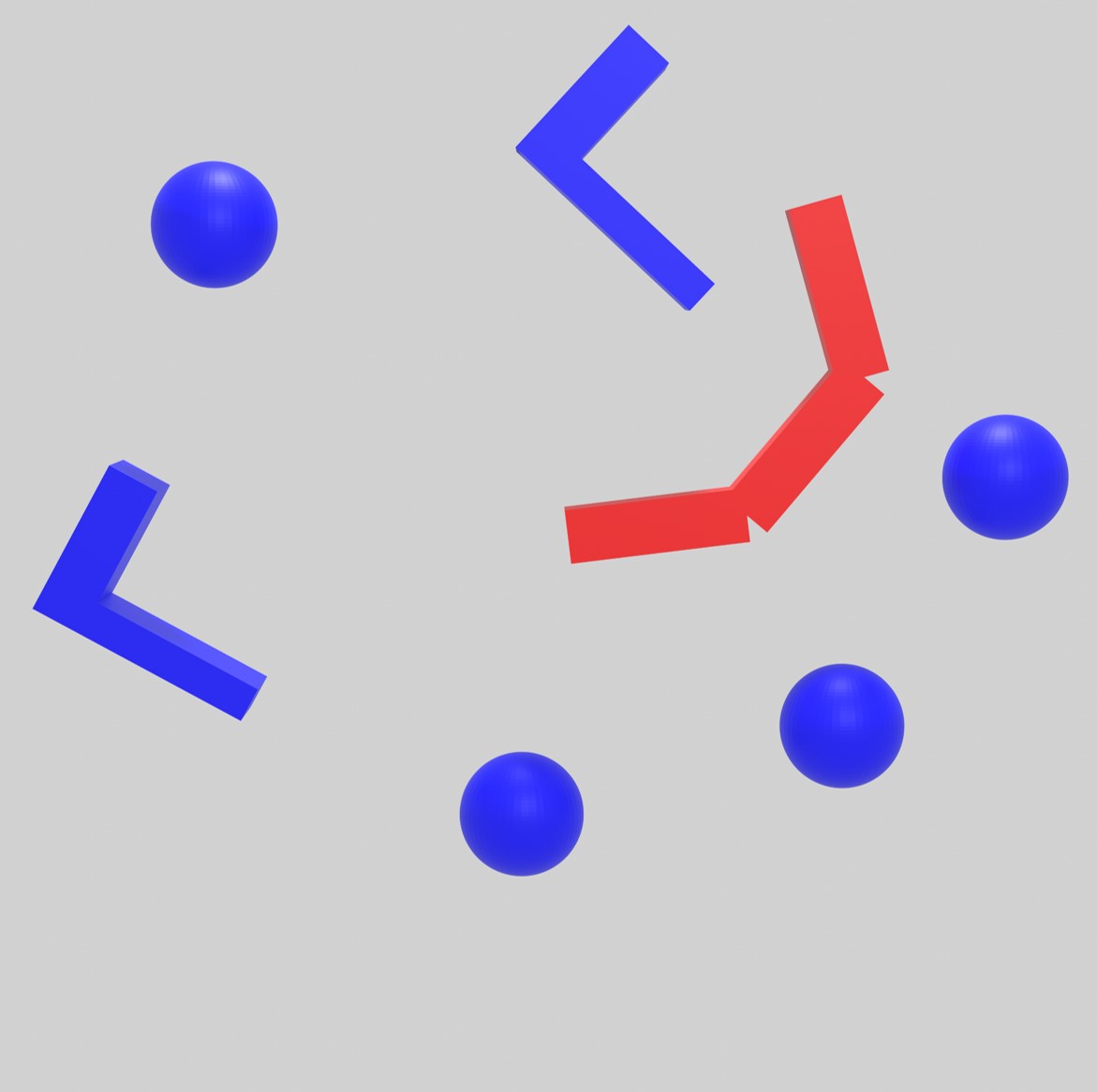}
			\caption{Constrained environment.}
			\label{fig:snake_robot}
		\end{subfigure}
		\hfill
		\begin{subfigure}[t]{.49\linewidth}
			\centering
			\includegraphics[width=\textwidth]{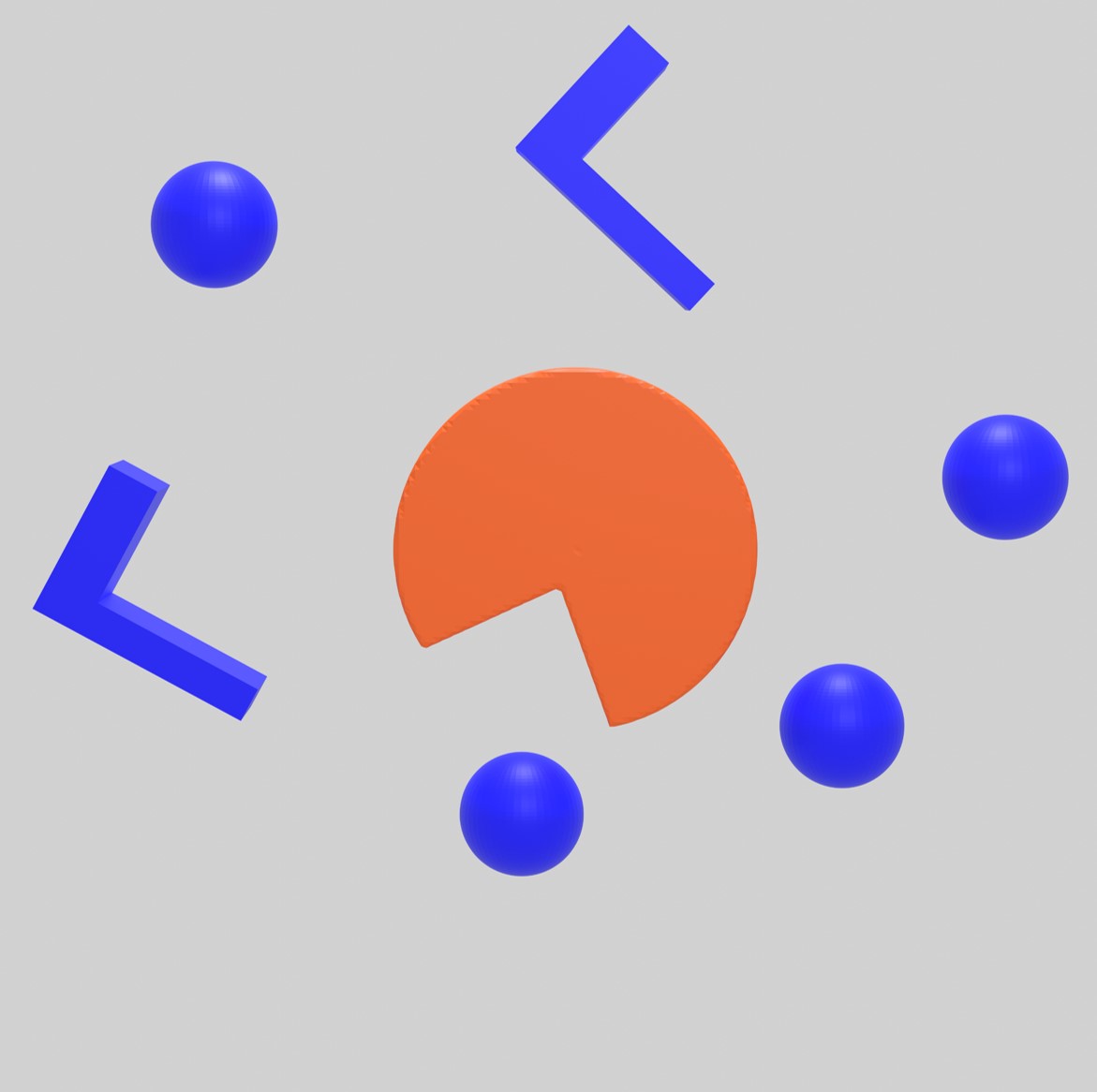}
			\caption{SV of the first link.}
			\label{fig:snake_sv_l1}
		\end{subfigure} \\
		\begin{subfigure}[t]{.49\linewidth}
			\centering
			\includegraphics[width=\textwidth]{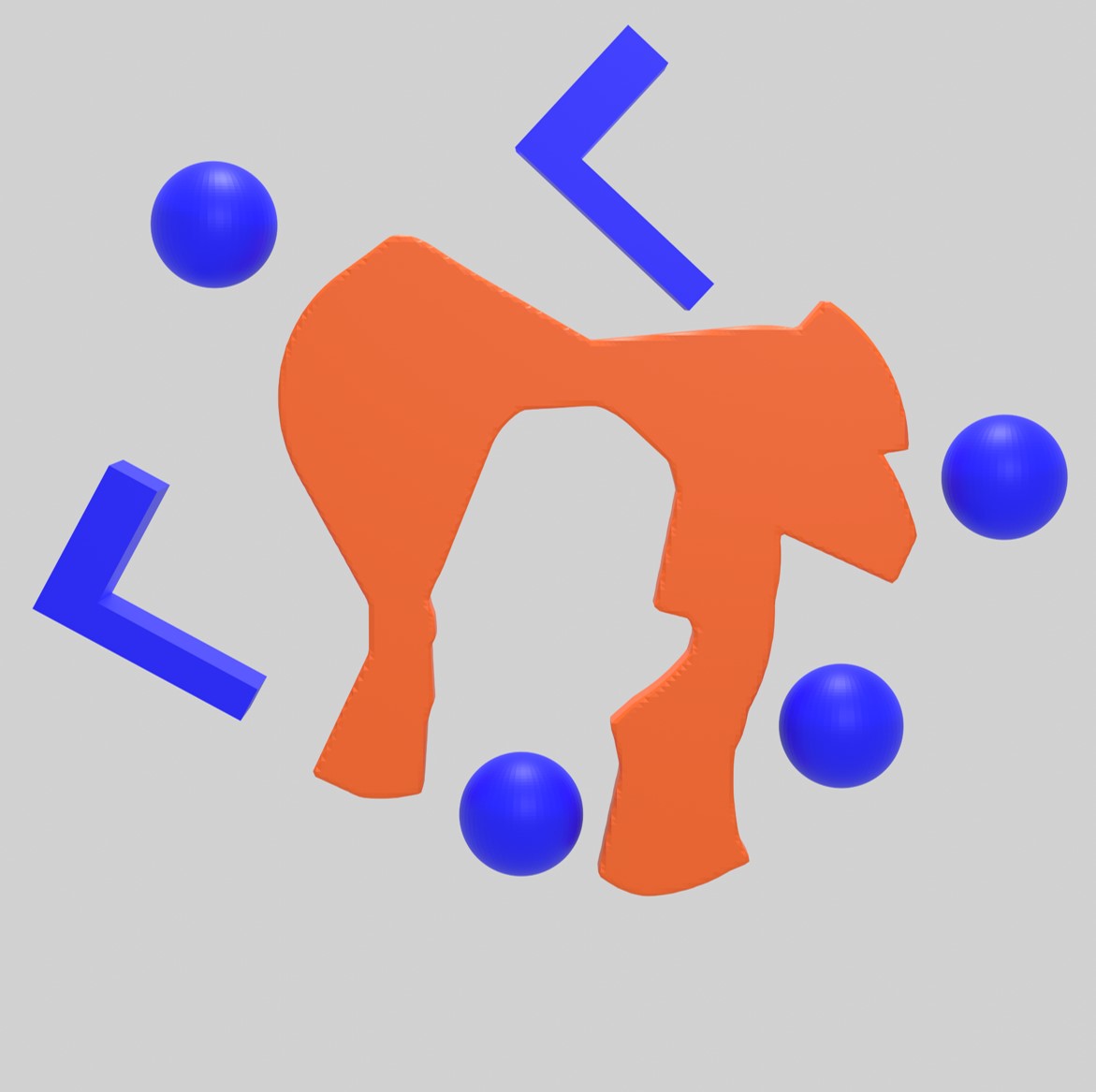}
			\caption{SV of the second link.}
			\label{fig:snake_sv_l2}
		\end{subfigure}
		\hfill
		\begin{subfigure}[t]{.49\linewidth}
			\centering
			\includegraphics[width=\textwidth]{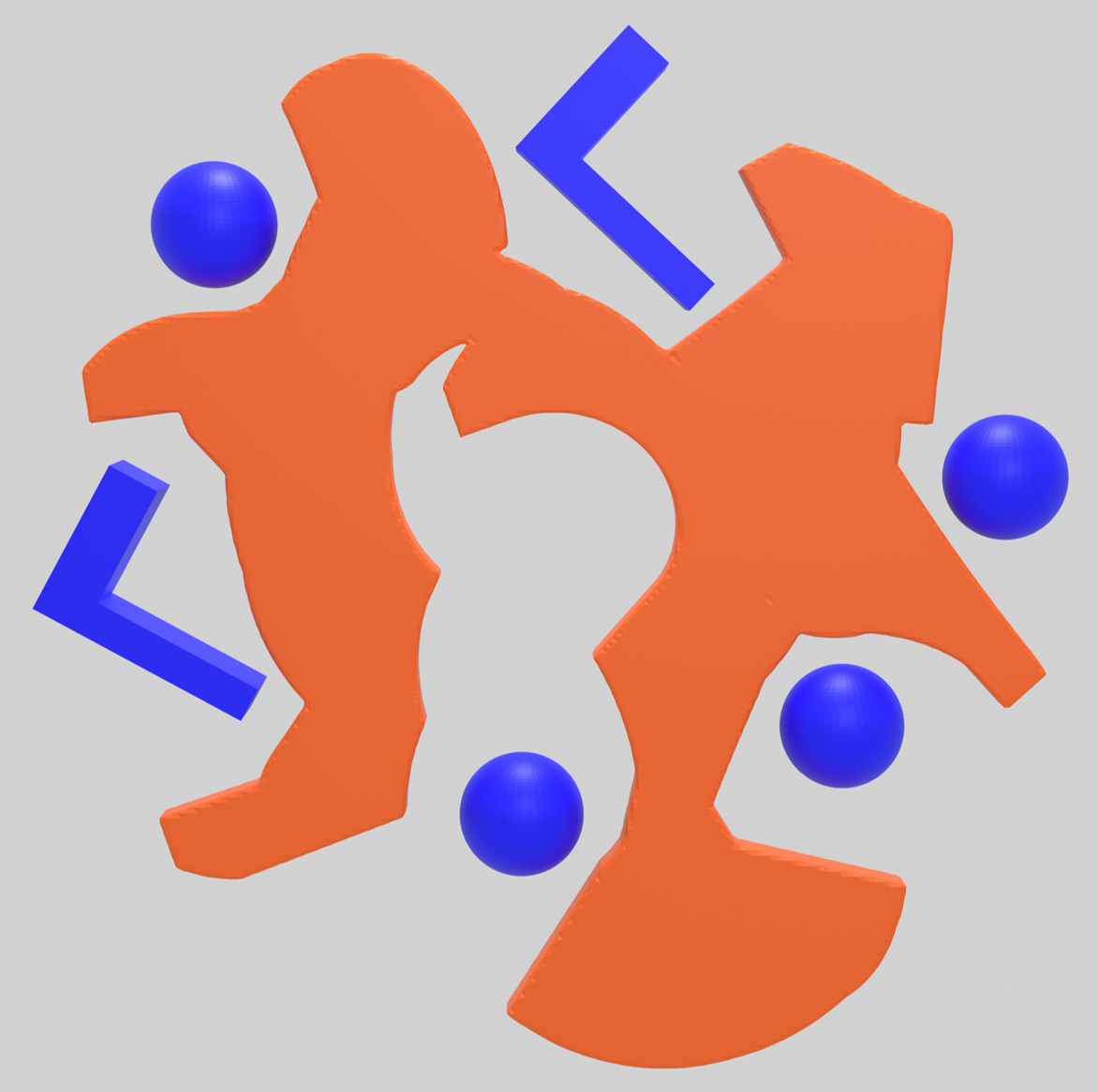}
			\caption{SV of the third link.}
			\label{fig:snake_sv_l3}
		\end{subfigure}		
	\end{minipage}
	\hfill
	\begin{minipage}{.49\linewidth}
		\begin{subfigure}[t]{\linewidth}
			\centering
			\includegraphics[width=\textwidth,height=9.24cm]{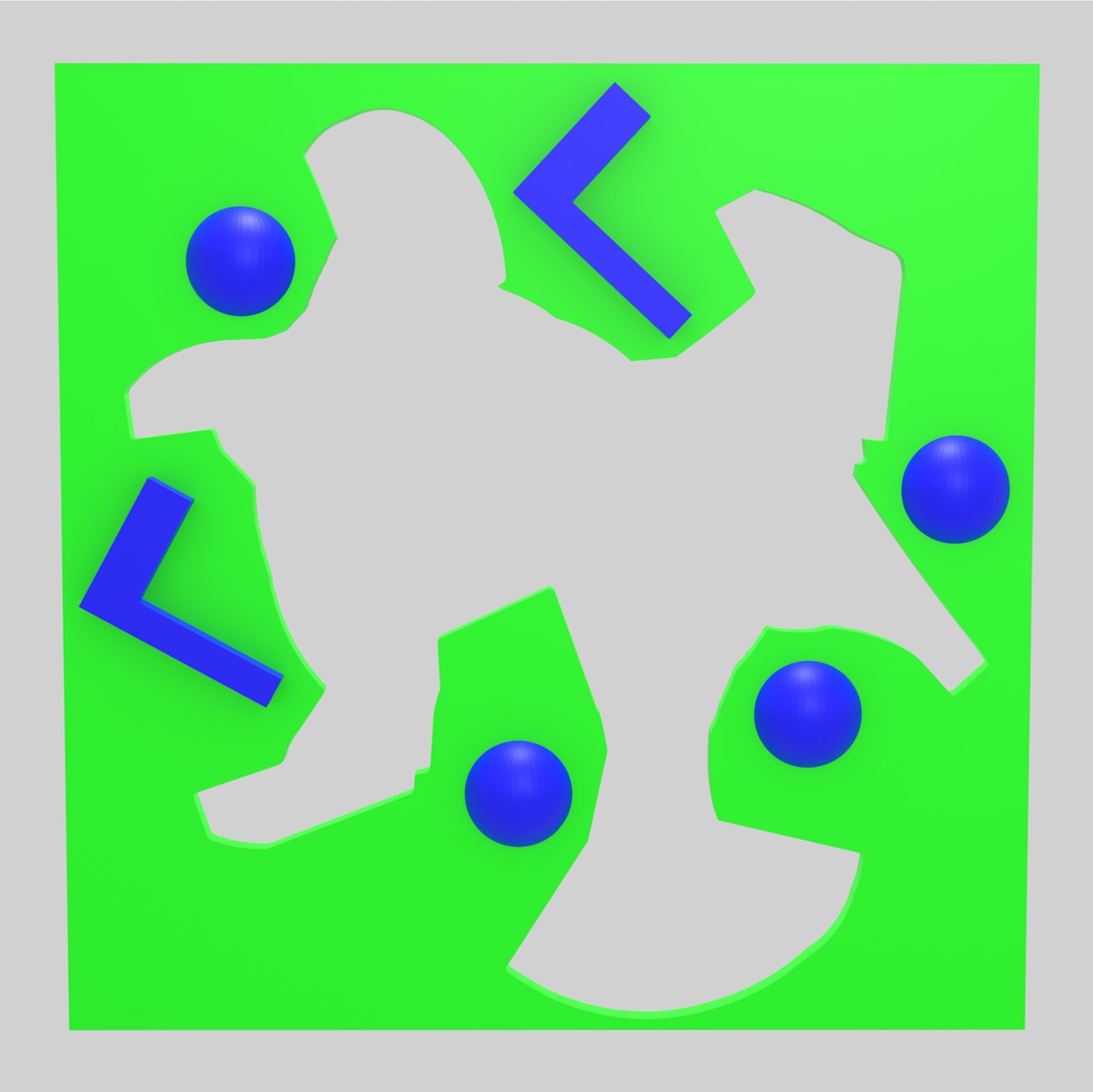}
			\caption{Obstacle representation (in green) with overlapping environment.}
			\label{fig:snake_obstacle_space}
		\end{subfigure}
	\end{minipage}
	\caption{Results obtained with step 2 and 4 of our proposed pipeline for a planar robot (in red) with three non-static links.}
	\label{fig:snake_experiment}
\end{figure*}

\Crefrange{fig:snake_sv_l1}{fig:snake_sv_l3} show the highly non-convex swept volumes associated with the three links of a planar robot, 
representing altogether the collision-free space discovered during the exploratory phase. 

\begin{figure}[t!]
	\centering
	\includegraphics[width=0.99\linewidth]{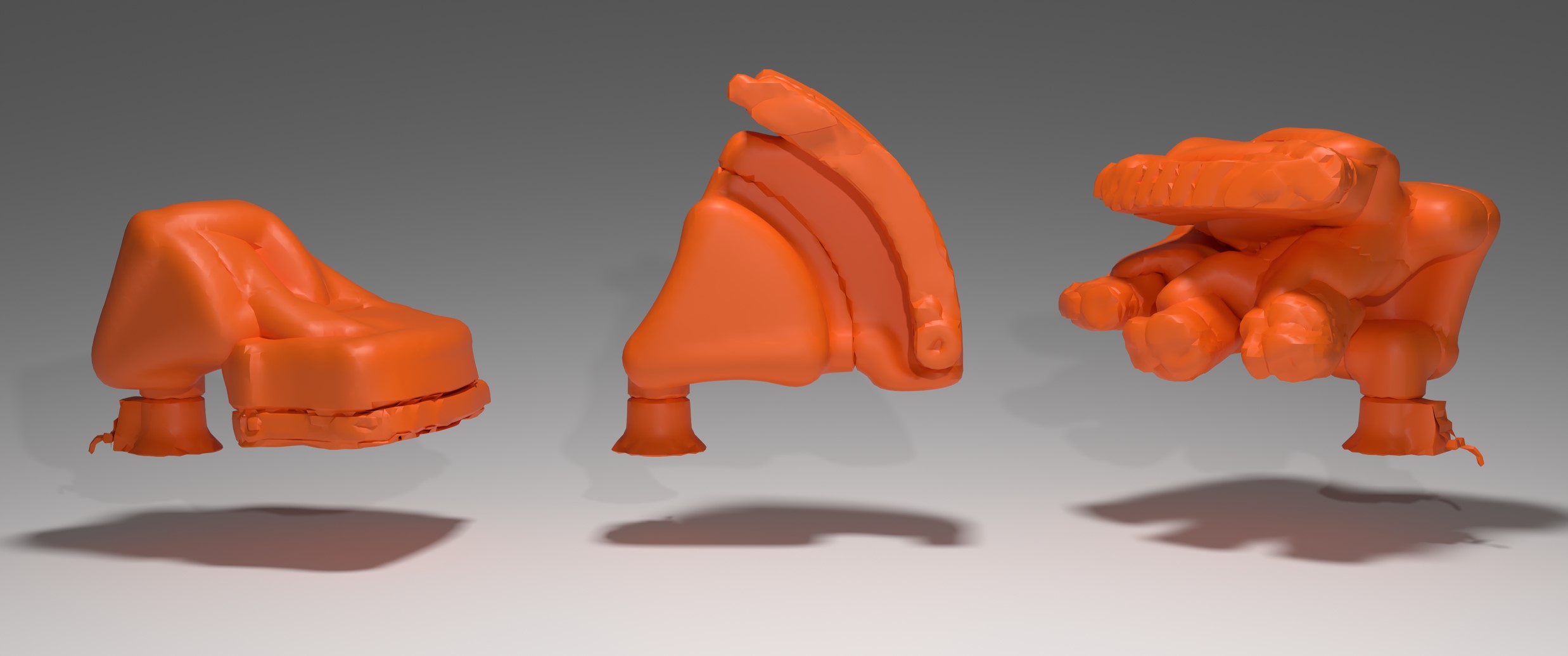}
	\caption{\new{
			Swept volumes of a KUKA LBR iisy 3 for three different motions: a rectangular (left) and semi-circular (middle) path for a gluing task, and the exploration of a shelf (right) for an inspection task.
	}}
	\label{fig:diverse_motions}
\end{figure}

\new{
	\refFig{fig:diverse_motions} illustrates examples of three-dimensional SVs of a KUKA LBR iisy manipulator performing three motions related to gluing and inspection tasks. 
	The mesh details including vertices, faces, and volume\footnote{\new{The volume of the SV meshes has been computed with the \textsc{3d print toolbox} of Blender.}}, 
	as well as the computation times, 
	are reported in \refTab{tab:results_preliminary_tests}.
}

\subsection{Volume Decimation (Optional)}
\label{subsec:volumeDecimation}
The usage of complex mesh representations in the context of computer graphics and 
3D modeling has encouraged research into new techniques
to (conservatively) simplify and reduce the size of a given mesh. 
This has resulted in the development of several decimation algorithms~\cite{cignoni1998comparison}.
These methods modify the 3D model by reducing the number of vertices, 
edges, 
and/or faces without (or conservatively) altering its overall shape and volume. 
This is achieved by identifying elements not strictly necessary to describe the object mesh.
Decimation algorithms iterate until a user-specified termination condition is met,
typically defined as a face reduction percentage.
In the following, 
the decimated volume of the $i$-th link SV is referred to as $\decimatedVolume{i}$.
This step is optional within our pipeline 
and often beneficial to reduce overall computation times (see \refSec{subsec:roleOfVolDeci}).

\subsection{Obstacle Representation}
\label{subsec:obstacleRepresentation}
Next, we compute a representation of the unexplored and (potentially) occupied volume $\volume{O}$.
It is obtained through
\begin{equation}
	\label{eq:obstacleSpace}
	\volume{O}=
	\left(
	\left(  
	\left(
	\volume{BV}
	~\backslash~
	\volume{1}
	\right)
	~\backslash~
	\volume{2}
	\right)
	~\backslash~
	\ldots~
	\right)
	~\backslash~
	\volume{\numLinks},
\end{equation}
where $\volume{BV}$ denotes a bounding volume covering the entire robot workspace.
An intuitive choice for the bounding volume in the case of a robot manipulator could be a sphere or cube parameterized according to the robot size.
The sequential operations in~\eqref{eq:obstacleSpace} are performed using the Boolean difference operator.
Initially defined in the Boolean algebra, 
Boolean operations have extended to computer graphics and 3D modeling.
The Boolean difference of two solid models 
$\mathcal{A},\mathcal{B}\subsetOfR{3}$ 
is defined as
$ 
	\mathcal{A}\,\backslash\,\mathcal{B} \coloneq
	\left\{
	\bm{x} \inR{3} 
	\mid 
	\bm{x} \in \mathcal{A}
	~\mathrm{and}~
	\bm{x} \notin \mathcal{B}
	\right\}.
$ 
Given $\numLinks$ swept volumes associated with the non-static robot links, 
the Boolean difference operation in~\eqref{eq:obstacleSpace} is performed $\numLinks$ times\footnote{The 
static base link of a fixed-base manipulator is not considered in~\eqref{eq:obstacleSpace}.}.
The resulting volume~$\volume{O}$ 
corresponds to the bounding volume~$\volume{BV}$
without a portion of the inner part.
Hence, 
$\volume{O}$~can be interpreted as a conservative mesh representation including all obstacles and unexplored areas in the confined workspace.
The robot is guaranteed to be collision-free as long as it does not penetrate~$\volume{O}$.

\refFig{fig:snake_obstacle_space} shows the unexplored and potentially occupied space 
obtained by considering the link swept volumes of the planar robot previously mentioned.

\subsection{Repeat (Optional)}
\label{subsec:repeatStep}
Suboptimal explorations can occur when the robot only partially sweeps the free workspace of interest.
Hence, 
unexplored areas will automatically be considered obstacles in the representation $\volume{O}$~obtained from~\eqref{eq:obstacleSpace},
negatively affecting the subsequent motion planning and control.
Therefore, 
after inspecting $\volume{O}$
(e.g., potentially using Augmented Reality glasses), 
the operator can perform additional exploration sessions (steps 1\,--\,3).
This way, 
the volume $\volume{O}$ will be further reduced, 
facilitating the collision-free motion planning.

\subsection{Discussion on Formal Guarantees}
Our approach is closely related to~\cite{Seidel2014}, 
describing a purely data-driven technique to environment modeling.
Given similar exploratory robot motions, 
a neural network for collision-free inverse kinematics is trained
and a graph representation in the task-space is built.
Each node in the graph represents a collision-free configuration.
Edges are introduced based on heuristics to connect neighboring nodes,
however, 
this process is not supported by any formal guarantee of a collision-free transition between nodes\footnote{Refer 
also to the figures 4 and 5 in~\cite{Seidel2014}.}.
Therefore, 
the authors suggest not to collect training data in proximity of the obstacles.
Furthermore, 
the method relies on a non-trivial distance metric that integrates both task- and joint-space information,
and requires the tuning of several hyper-parameters.
Together, 
the neural network and graph, 
implicitly form an environment model, 
which is utilized for autonomous motion planning.
It is, 
however, 
not suitable for reactive control.

In contrast, 
our approach explicitly generates an accurate environment model represented as a mesh, 
thus providing formal guarantees: 
as long as the robot moves within the previously explored space, 
collision-free motions are guaranteed.
This is achieved through algorithms for computing SVs and boolean operations that produce correct or conservative results.
Another advantage of our mesh representation is that it can be integrated with state-of-the-art optimization techniques 
to ensure proven collision-free trajectory planning and control, 
which is crucial for safety-critical industrial applications.

\subsection{Limitations}
\new{
	Our method is explicitly designed for deployment in static robot cell environments (Assumption A1). 
	However, 
	many real-world applications involve dynamic environments with moving obstacles and objects (e.g., on a conveyor),
	which can lead to unintended robot collisions. 
	To automate processes in these environments, 
	we suggest integrating the environment model generated by our approach with real-time data from additional sensors 
	to detect and track dynamic obstacles and/or objects.
	On the other hand, 
	it is worth noticing that the assumptions A2 and A3 generally hold in industrial applications.
	
		
	
	Currently, 
	our software implementation is not capable of real-time processing. 
	Therefore, 
	it is not possible to provide visual feedback of the growing robot swept volume during the exploration phase, 
	though such feedback would be beneficial for the operator. 
	We expect that future advancements in swept volume computation will help address this limitation. 
}

\section{Experiment}
\label{sec:experiments}
\noindent We validate the proposed \new{pipeline} within a pick-and-place scenario, confirming its effectiveness.

\begin{table}[t!]
	\caption{Swept volume mesh details and computation time (ct) for a kuka lbr iisy 3 performing three different motions.
	}
	\centering
	\label{tab:results_preliminary_tests}
	\begin{tabularx}{0.49\textwidth}{M{6.05em}|C|C|C|C}
		\hline
		Motion & Vertices & Faces & Volume & CT \\
		\hline
		rect. path & $19399$ & $38810$ & $\volMeasure{0.43}$ & $\timeInSec{24.53}$ \\
		\hline
		semi-circ. path & $13093$ & $26182$ & $\volMeasure{0.29}$ & $\timeInSec{14.35}$ \\
		\hline
		shelf expl. & $26724$ & $53502$ & $\volMeasure{0.69}$ & $\timeInSec{57.90}$ \\
		\hline
	\end{tabularx}
\end{table}

\subsection{Software Libraries}
In this work, 
the SV computation~\eqref{eq:sweptVol} is based on the algorithm described in~\cite{sellan2021swept}.
It is implemented within the \textsc{gpytoolbox} 
library\footnote{S. Sellán, O. Stein et al., “gptyoolbox: 
	A python geometry processing toolbox,” 2023, \url{https://gpytoolbox.org/}.}, 
which leverages the \textsc{libigl} 
library\footnote{A. Jacobson, D. Panozzo et al., “libigl: 
	A simple C++ geometry processing library,” 2018, \url{https://libigl.github.io/}.}.
The algorithm inputs are the triangle mesh of the solid of interest and its discretized sequence of poses,
resulting in a high-quality 3D mesh.

We selected the \textsc{visualization toolkit}~\cite{vtkBook} for the decimation algorithm.
The input is a triangle mesh, 
and the parameters defining the decimation process 
(e.g., the target percentage of the triangle reduction, 
the maximum allowed error, and whether the mesh topology should be preserved).
It returns a triangle mesh.

Finally, 
we used the software library \textsc{libigl} 
for the Boolean difference operation, 
which is based on~\cite{zhou2016mesh} and 
\textsc{cgal} library\footnote{"CGAL, Computational Geometry Algorithms Library," \url{https://www.cgal.org/}.}.
The algorithm inputs are two triangle meshes, 
provided as a collection of vertices and faces.

\begin{figure}[t!]
	\centering
	\includegraphics[width=0.99\linewidth]{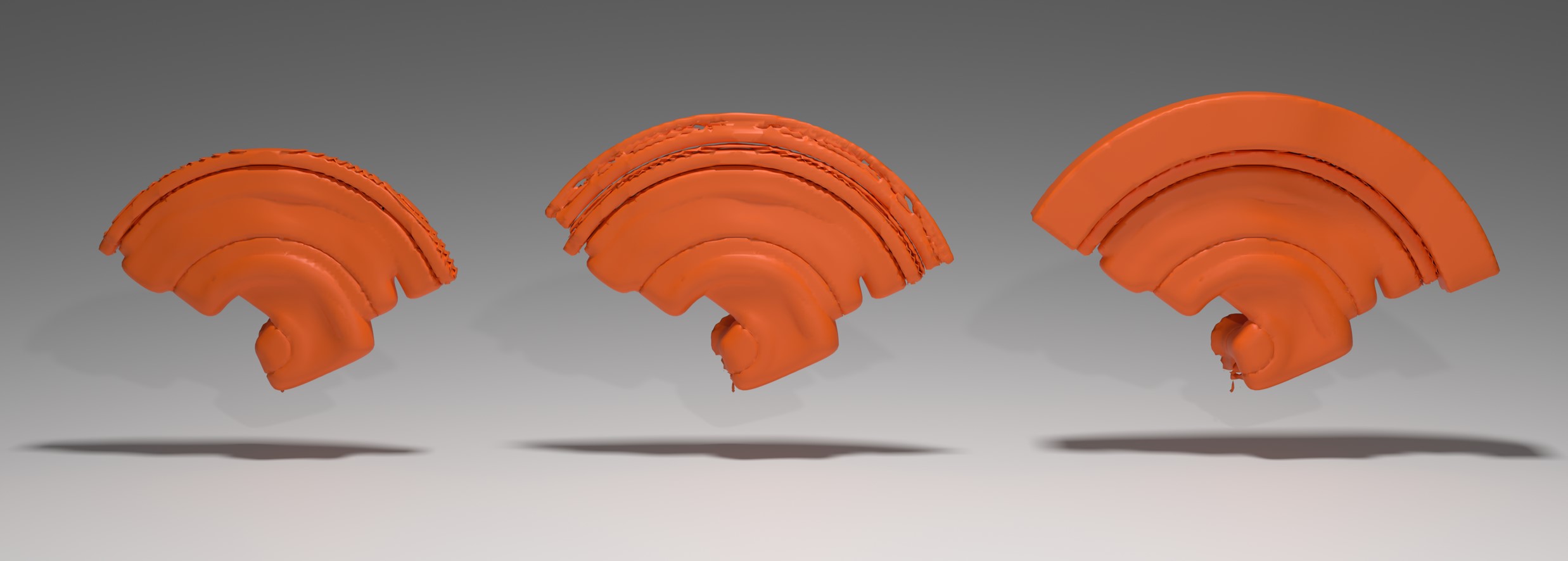}
	\caption{KUKA LBR iisy swept volume of a simple trajectory: 
		robot as it is (left), 
		with a parallel gripper (middle), 
		and with a cube-shaped exploration tool (right).}
	\label{fig:sweptvols_comparison}
\end{figure}

\subsection{Hardware Setup}
\begin{figure}[b!]
	\centering
	\includegraphics[width=0.99\linewidth]{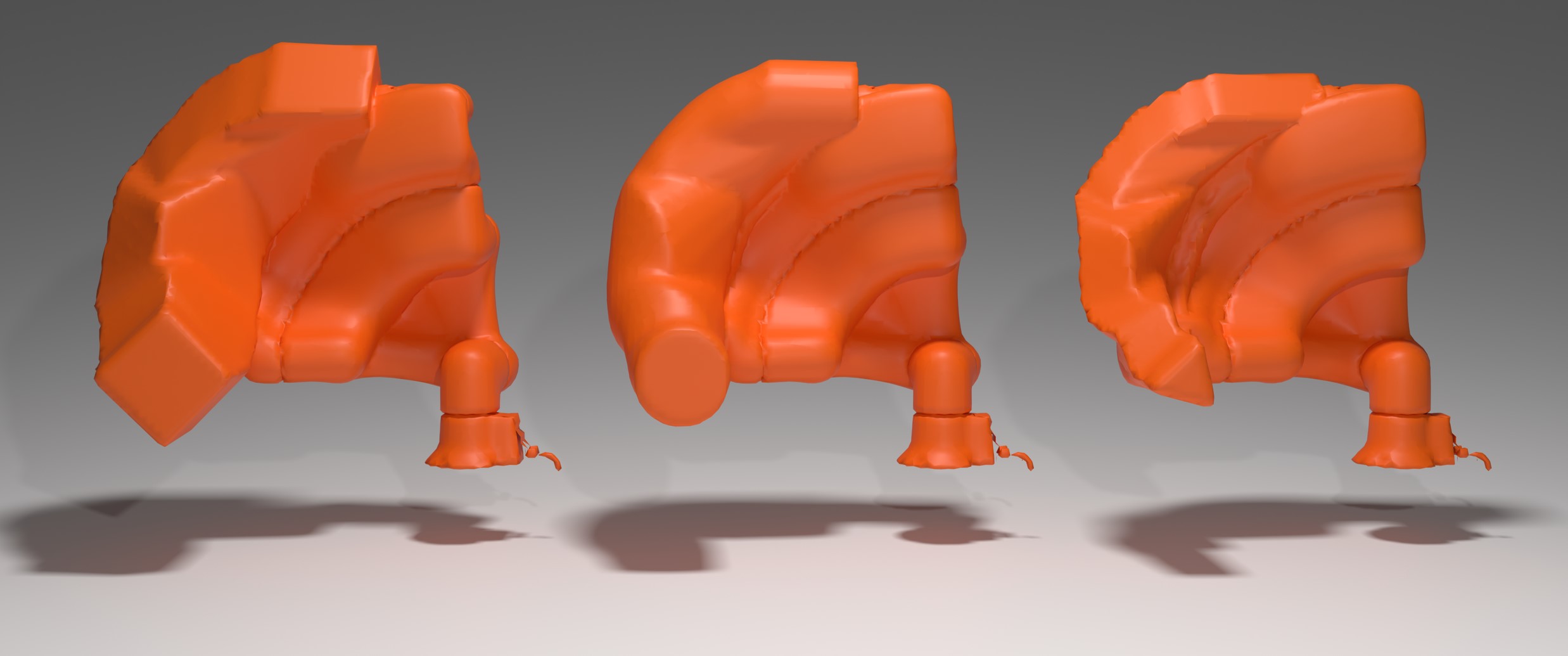}
	\caption{\new{
			Swept volumes of a KUKA LBR iisy 3 executing a simple trajectory with three distinct exploration tool geometries: a cube (left), a cylinder (middle), and a triangular prism (right).
			}}
	\label{fig:multiple_exploration_tools}
\end{figure}

The experimental platform is a KUKA LBR iisy 3 R760 with $\numLinks=6$ non-static links.
\new{The robot} is equipped with a SCHUNK change system (FWA series), 
enabling a quick switch between the exploration tool and the gripper required for the task.
The latter is a SCHUNK gripper (GEI FWA-50 series) with parallel 3D-printed fingertips.  Overall, it approximately measures $\distInCm{\left( 7 \times 7 \times 12 \right)}$.
We designed a cube-shaped exploration tool to speed up the exploration phase 
and avoid high computation times or numerical instabilities. 
\refFig{fig:exploration_frames} shows the exploration tool, 
a lightweight cardboard box of $\distInCm{\left( 20 \times 20 \times 20 \right)}$, which encloses the gripper entirely.
\refFig{fig:sweptvols_comparison} highlights the advantages of using this tool
by comparing the robot SV mesh for a \new{planar} trajectory in three different scenarios: 
the robot as it is, 
with a parallel gripper, 
and with the exploration tool.
The SV mesh generated with the gripper shows imperfections such as holes and irregularities. 
In contrast, 
the use of the exploration tool results in a smoother and more expansive SV.
This underlines our secondary contribution. 
\new{
	\refFig{fig:multiple_exploration_tools} displays additional SV examples involving three distinct exploration tool geometries within a simple trajectory.	
}
Utilizing Robot Operating System (ROS)~2 
to exchange data,
all computations are performed on a laptop system with an Intel Core i7-12800H (2.4 GHz) CPU and 32 GB of RAM.


\subsection{Constrained Robot Cell Environment}
\begin{figure}[!t]
	\centering
	\includegraphics[width=0.99\linewidth]{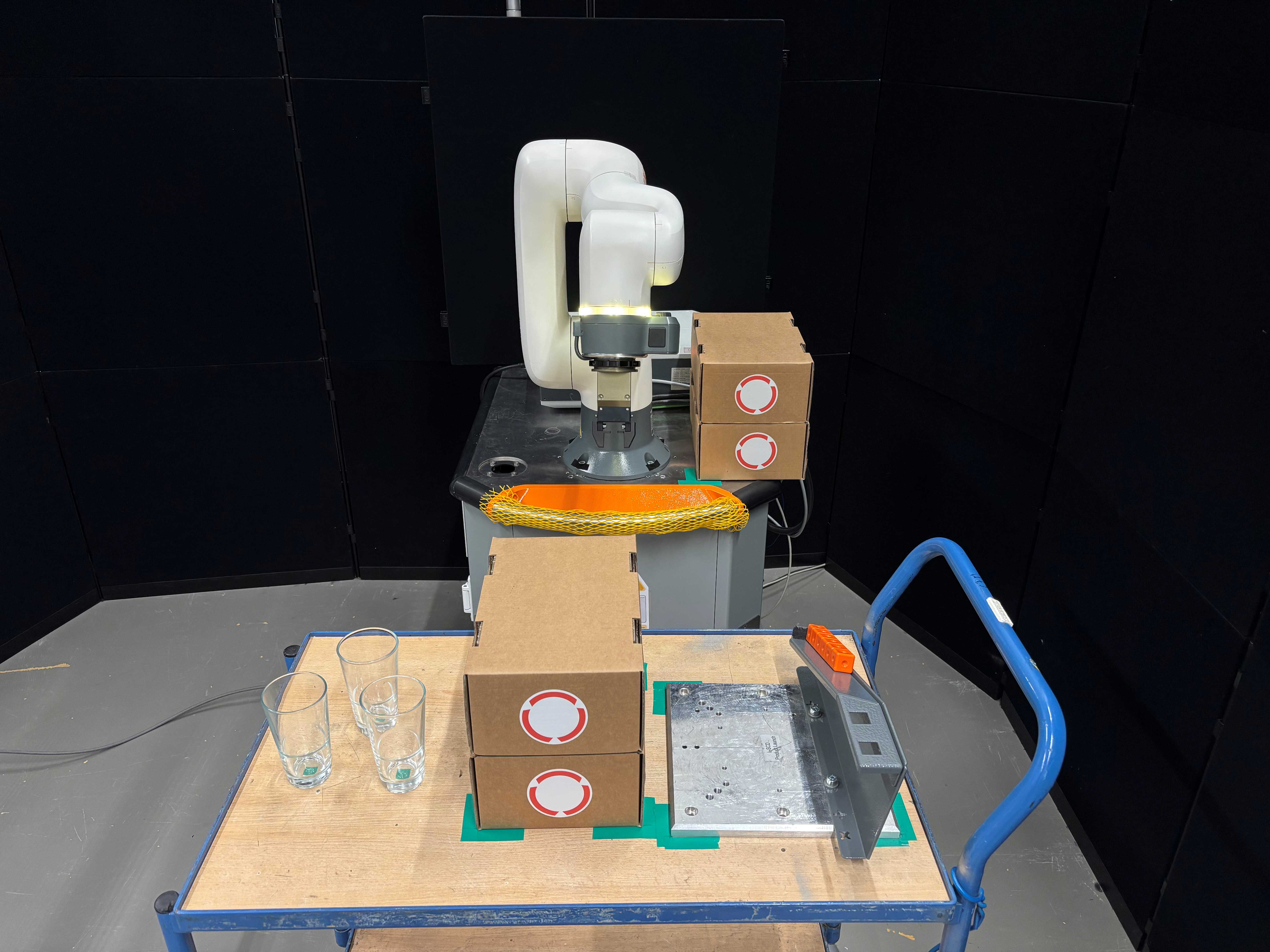}
	\caption{Constrained cell with a KUKA LBR iisy performing a pick-and-place task. 
	The robot must pick the orange blocks from the ramp and place them into the three glasses 
	by exploiting the proposed obstacle representation.}
	\label{fig:demonstrator}
\end{figure}

\refFig{fig:demonstrator} shows an industrial cart positioned in front of the robot,
with three glasses and a ramp holding six orange blocks. 
The cell contains four additional obstacles: 
two boxes between the ramp and the glasses and two boxes next to the robot, 
limiting its elbow motions. 
While operating, 
the robot must avoid self-collisions and collisions with the environment.

\subsection{Pick-and-Place Task}
\label{subsec:pickAndPlaceTask}
The robot must pick the orange blocks from the ramp and place them into the glasses. 
\new
{
	Whenever the block at the bottom of the ramp is removed, 
	the remaining blocks shift downward, 
	preserving a constant picking location.
}
This process is repeated six times until the ramp is empty. 
The first three blocks go into different glasses, 
and the next three follow the same order. 
In the end, 
each glass contains two blocks.

\subsubsection{Workspace Exploration}
The operator explores the collision-free space with the collaborative robot, 
equipped with the exploration tool, 
utilizing the hand guidance mode proposed in~\cite{osorio2019physical}.
We record the trajectory at $25$ Hz 
and collect $4009$ joint configurations in less than three minutes.
\refFig{fig:exploration_frames} shows selected video frames of the supplementary material.

\begin{figure*}[!t]
	\centering
	\includegraphics[width=0.99\linewidth]{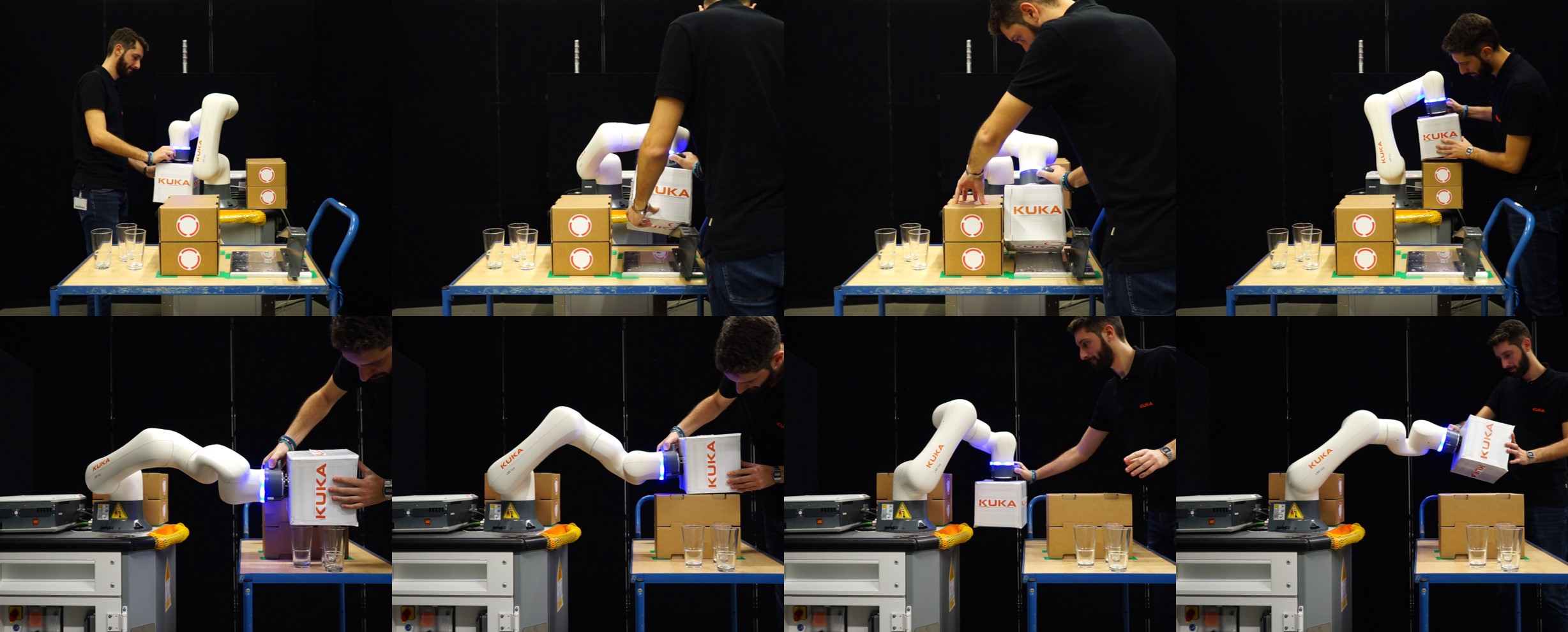}
	\caption{Exploration phase through hand guidance of a KUKA LBR iisy equipped 
	with a cube-shaped exploration tool: 
	front view (top row) and side view (bottom row).
	The corresponding swept volume is shown in~\refFig{fig:sv_pick_and_place}.}
	\label{fig:exploration_frames}
\end{figure*}

\subsubsection{Robot Link Swept Volumes and Decimation}
The six swept volumes of the non-static links are decimated and jointly visualized in \refFig{fig:sv_pick_and_place} together with the static robot base.
The volume decimation step has reduced both the number of vertices and faces by $63.64 \,\%$.

\subsubsection{Obstacle Representation}
We represent the bounding volume $\volume{BV}$ as a cube whose dimensions have been chosen according to the maximum robot length.
The mesh of the obstacle representation volume $\volume{O}$ is obtained by iteratively subtracting the mesh representing the link swept volumes $\decimatedVolume{i}$ from the mesh of the bounding volume $\volume{BV}$.

\subsubsection{Motion Planning and Control}
Thanks to the change system,
once the exploratory phase has been performed, 
the gripper quickly replaces the exploration tool. 
The pick-and-place task is modeled as a simple finite state machine,
specifying the pick-and-place poses through the hand guidance mode.
Moreover, 
since a collision is likely detected when the robot picks up an orange block, 
we record an additional \emph{pre-pick} joint configuration $16$ cm above the ramp.

\new{
	Operators can flexibly select from a range of motion planning frameworks, 
	including QP-based methods, AI-based planners, global planners, or reactive strategies, 
	depending on the demands of the particular application.
	In this experiment, 
	we rely on the framework presented in \cite{osorio2020unilateral} and the \textsc{flexible collision library (fcl)}~\cite{pan2012fcl} to generate optimal collision-free trajectories through fast collision checks.
	Potential collisions are evaluated given the mesh of the robot, gripper, 
	and the environment represented as $\volume{O}$.
	The motion framework also considers robot self-collisions.
}

\subsubsection{Task Execution}
The robot executes the planned trajectories without collisions, 
moving consistently within the previously explored space. 
For further details, 
refer also to the video in the supplementary material.

\subsection{Execution Time Analysis}
\label{subsec:exTimeAnalysis}
A thorough exploration (performed by the first author) of the free space surrounding the robot took $\timeInSec{162.27}$ ($\approx\timeInMin{2.70}$).
\refTab{tab:exTimesPickAndPlace} reports the execution times of each pipeline step
(see \refSec{subsec:pickAndPlaceTask}) averaged over ten independent pipeline executions. 
From a computational perspective, 
the most time-consuming operation is the computation of the robot link swept volumes, 
which takes $\timeInSec{185.81}$ ($\approx\timeInMin{3.1}$). 
On the other hand, 
the computation of the volume decimation is almost negligible, 
taking only a few seconds.
\new{
	Overall, 
	the manual robot workspace exploration and the automatic generation of the environment model~$\volume{O}$ took $\timeInSec{368.49}$ ($\approx\timeInMin{6.14}$).
}
Additionally, 
programming the particular application \new{required approximately} $\timeInMin{1.5}$ to 
record the home robot configuration\new{, set the pick-and-place poses, and generate a collision-free trajectory.}
In summary, 
the robot operator spent $\timeInMin{7.64}$ to model the robot cell and setup the collision-free robot program. 

\begin{table}[t!]
	\caption{Execution times in seconds averaged over ten independent executions
			 of the pipeline steps for a pick-and-place task with a kuka lbr iisy collaborative robot.
			 }
	\centering
	\label{tab:exTimesPickAndPlace}
	\begin{tabularx}{0.49\textwidth}{C|C|C|C|C}
			\hline
			\multirow{2}{*}{Exploration} & Swept & Volume & Obstacle & \multirow{2}{*}{Total}\\ 
			& Volume & Decimation & Repr. & \\
			\hline
			$\timeInSec{162.27}$ & $\timeInSec{185.81}$ & $\timeInSec{2.16}$ & $\timeInSec{18.25}$ & $\timeInSec{368.49}$ \\
			\hline
		\end{tabularx}
\end{table}

\subsection{The Role of the Volume Decimation}
\label{subsec:roleOfVolDeci}
In this ablation study, 
we evaluate the role of the volume decimation step in the pipeline from a computational load point of view. 
Therefore, 
we have repeated the execution time analysis conducted in \refSec{subsec:exTimeAnalysis} without performing the volume decimation step. 
The obstacle representation step works directly with the SV meshes $\volume{i}$ produced by the previous step.
The execution times to obtain the obstacle representation increase from $\timeInSec{18.25}$ to $\timeInSec{37.40}$. 
Hence, 
the optional and almost costless execution of $\timeInSec{2.16}$ for the volume decimation step implies a reduction of $\timeInSec{16.99}$ ($\approx 4.41\,\%$) on the whole execution time. 
It is worth noticing that computational optimization may be significantly higher in the case of longer explorations or a more complex SV mesh.

\section{Conclusion}
\label{sec:conclusion}
\noindent This letter proposes a novel data-driven and robot-agnostic approach to modeling obstacles within a cluttered robot cell. 
The method does not rely on additional external sensors, 
making the environment modeling process cost-effective and immediate. 
It supports novice users, who can gather the necessary data by hand guiding the robot.
After performing exploratory robot motions for few minutes,
the unexplored and potentially occupied space is modeled by leveraging the robot's kinematic structure and the swept volume of the non-static links.
Our method is capable of effectively managing clustered or heterogeneously distributed data.
We obtain a triangular mesh, 
which is finally used to plan and execute collision-free trajectories safely.
Showcasing the method's potential to streamline industrial processes,
we validated the intuitive interface in a pick-and-place scenario.
Our execution time analysis highlighted that the user can model a robot cell and perform a task in less than eight minutes. 
Moreover, 
the ablation study showed the beneficial role of the optional volume decimation (step 3), 
which further optimizes the computational efficiency almost at no cost. 

Future work involves integrating an autonomous exploration mode, 
where the robot changes its direction of motion upon contact detection.
Furthermore, 
we would like to evaluate less accurate but faster AI-accelerated techniques similar to \cite{baxter2020deep,joho2024neural}
for visualizing the swept volume in real-time during an exploratory session (e.g., utilizing immersive augmented reality hardware).
\new{
	Finally, 
	a user study demonstrating the programming efficiency compared to conventional methods could offer valuable insights to the community.
}

\section{Acknowledgments}
\noindent This work was partly supported by KUKA Deutschland GmbH
and the state of Bavaria through the OPERA project DIK-2107-0004/DIK0374/01.

\balance

\bibliographystyle{IEEEtran}
\bibliography{bibliography.bib}

\end{document}